\documentclass[11pt,a4paper]{article}
\usepackage[hyperref]{acl2019}
\usepackage{times}
\usepackage{latexsym}
\usepackage{comment}
\usepackage{amsmath}
\mathchardef\mhyphen="2D
\usepackage{graphicx}
\graphicspath{ {./} }
\usepackage{float}
\usepackage[linguistics]{forest}
\usepackage{xcolor}

\usepackage{url}

\usepackage{url}

\aclfinalcopy % Uncomment this line for the final submission
\def\aclpaperid{1592} %  Enter the acl Paper ID here

\title{Graph based Neural Networks for Event Factuality Prediction \\using Syntactic and Semantic Structures}

\author{Amir Pouran Ben Veyseh, Thien Huu Nguyen and Dejing Dou  \\
  Department of Computer and Information Science\\
  University of Oregon\\
  Eugene, OR 97403, USA \\
  \texttt{\{apouranb, thien, dou\}}\texttt{@cs.uoregon.edu} \\
}

\date{}

\begin{document}
\maketitle
\begin{abstract}
Event factuality prediction (EFP) is the task of assessing the degree to which an event mentioned in a sentence has happened. For this task, both syntactic and semantic information are crucial to identify the important context words. The previous work for EFP has only combined these information in a simple way that cannot fully exploit their coordination. In this work, we introduce a novel graph-based neural network for EFP that can integrate the semantic and syntactic information more effectively. Our experiments demonstrate the advantage of the proposed model for EFP.
%, leading to the state-of-the-art results on different public datasets.
\end{abstract}

%connect the anchor words of the event mentions to the appropriate context words in the sentences

\section{Introduction}

Events are often presented in sentences via the indication of anchor/trigger words (i.e., the main words to evoke the events, called event mentions) \cite{Nguyen:16a}. Event mentions can appear with varying degrees of uncertainty/factuality to reflect the intent of the writers. In order for the event mentions to be useful (i.e., for knowledge extraction tasks), it is important to determine their factual certainty so the actual event mentions can be retrieved (i.e., the event factuality prediction problem (EFP)). In this work, we focus on the recent regression formulation of EFP that aims to predict a real score in the range of [-3,+3] to quantify the occurrence possibility of a given event mention \cite{stanovsky2017integrating,Rudinger}. This provides more meaningful information for the downstream tasks than the classification formulation of EFP \cite{lee2015event}. For instance, the word ``{\it left}'' in the sentence ``{\it She left yesterday.}'' would express an event that certainly happened (i.e., corresponding to a score of +3 in the benchmark datasets) while the event mention associated with ``{\it leave}'' in the sentence ``{\it She forgot to leave yesterday.}'' would certainly not happen (i.e., a score of -3).

% EFP is a challenging problem as different context words might jointly participate to reveal the factuality of the event mentions (i.e., the cue words), possibly located at different parts of the sentences and scattered far away from the anchor words of the events. For a model to be able to correctly identify the factuality, it should be able to locate the cue words for the event mention and appropriately attend to those context words and the anchor word to make the final judgment.

%determiners, adverbs

\begin{figure}
\centering
\begin{forest}
[go
  [I]
  [will]
  [seeing
    [after]
    [treatment
      [the]
      [others
        [of]]]]
  [need
   [when]
   [I]
   [care
     [medical]]]
  [back]
]
\end{forest}
\caption{The dependency tree of the sentence ``{\it I will, after seeing the treatment of others, go back when I need medical care.}''.} \label{parse}
\end{figure}
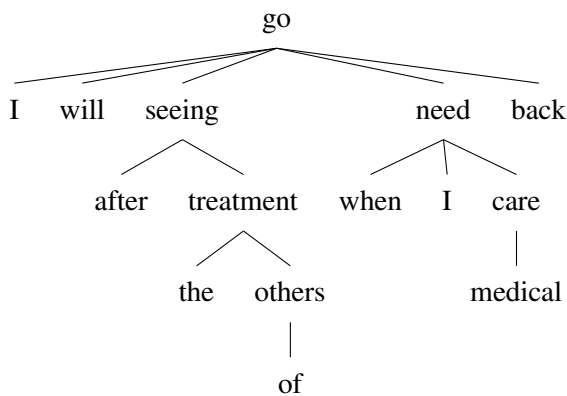

% \begin{figure}
% \centering
% \begin{forest}
% [\textcolor{green}{born}
%   [\textcolor{green}{president}
%     [new]
%     [U.S.]
%     [\textcolor{red}{Trump}
%         [\textcolor{red}{Donald}]]]
%   [was]
%   [14
%     [on]]  
%   [June
%     [1946]]
%   [\textcolor{green}{city}
%     [at]
%     [\textcolor{red}{New York}]]
% ]
% \end{forest}
% \caption{The dependency tree of the sentence ``{\it I will, after seeing the treatment of others, go back when I need medical care.}''.} \label{parse}
% \end{figure}

EFP is a challenging problem as different context words might jointly participate to reveal the factuality of the event mentions (i.e., the cue words), possibly located at different parts of the sentences and scattered far away from the anchor words of the events. There are two major mechanisms that can help the models to identify the cue words and link them to the anchor words, i.e., the syntactic trees (i.e., the dependency trees) and the semantic information \cite{Rudinger}. For the syntactic trees, they can connect the anchor words to the functional words (i.e., negation, modal auxiliaries) that are far away, but convey important information to affect the factuality of the event mentions. For instance, the dependency tree of the sentence ``{\it I will, after seeing the treatment of others, go back when I need medical care.}'' will be helpful to directly link the anchor word ``{\it go}'' to the modal auxiliary ``{\it will}'' to successfully predict the non-factuality of the event mention. Regarding the semantic information, the meaning of the some important context words in the sentences can contribute significantly to the factuality of an event mention. For example, in the sentence ``{\it Knight lied when he said I went to the ranch.}'', the meaning represented by the cue word ``{\it lied}'' is crucial to classify the event mention associated with the anchor word ``{\it went}'' as {\it non-factual}. The meaning of such cue words and their interactions with the anchor words can be captured via their distributed representations (i.e., with word embeddings and long-short term memory networks (LSTM)) \cite{Rudinger}.

% via the head-modifier relations between words so the models can successfully predict the non-factuality of the corresponding event mention.

The current state-of-the-art approach for EFP has involved deep learning models \cite{Rudinger} that examine both syntactic and semantic information in the modeling process. However, in these models, the syntactic and semantic information are only employed separately in the different deep learning architectures to generate syntactic and semantic representations. Such representations are only concatenated in the final stage to perform the factuality prediction. A major problem with this approach occurs in the event mentions when the syntactic and semantic information cannot identify the important structures for EFP individually (i.e., by itself). In such cases, both the syntactic and semantic representations from the separate deep learning models would be noisy and/or insufficient, causing the poor quality of their simple combination for EFP. For instance, consider the previous example with the anchor word ``{\it go}'': ``{\it I will, after seeing the treatment of others, go back when I need medical care.}''. On the one hand, while syntactic information (i.e., the dependency tree) can directly connect ``{\it will}'' to ``{\it go}'', it will also promote some noisy words (i.e., ``{\it back}'') at the same time due to the direct links (see the dependency tree in Figure \ref{parse}). On the other hand, while deep learning models with the sequential structure can help to downgrade the noisy words (i.e., ``{\it back}'') based on the semantic importance and the close distance with ``{\it go}'', these models will struggle to capture ``{\it will}'' for the factuality of ``{\it go}'' due to their long distance.

% From this example, we also see that the syntactic and semantic information can complement each other to both promote the important context words and blur the irrelevant words. Consequently, we argue that the syntactic and semantic information should be allowed to interact earlier in the modeling process to produce more effective representations for EFP. In particular, we propose a novel method to integrate syntactic and semantic structures of the sentences based on the graph convolutional neural networks (GCN) \cite{kipf2016semi} for EFP. The modeling of GCNs involves affinity matrices to quantify the connection strength between pairs of words, thus facilitating the integration of syntactic and semantic information. In the proposed model, the semantic structures of the sentences are induced from the distributed representations of the words obtained from LSTMs. Such learned semantic structures are then linearly integrated with the syntactic structures of the dependency trees, resulting in new structures for the sentences that weight the context words according to their semantic and syntactic importance for factuality prediction. The extensive experiments show that the proposed model is very effective for EFP.

From this example, we also see that the syntactic and semantic information can complement each other to both promote the important context words and blur the irrelevant words. Consequently, we argue that the syntactic and semantic information should be allowed to interact earlier in the modeling process to produce more effective representations for EFP. In particular, we propose a novel method to integrate syntactic and semantic structures of the sentences based on the graph convolutional neural networks (GCN) \cite{kipf2016semi} for EFP. The modeling of GCNs involves affinity matrices to quantify the connection strength between pairs of words, thus facilitating the integration of syntactic and semantic information. In the proposed model, the semantic affinity matrices of the sentences are induced from Long Short-Term Memory networks (LSTM) that are then linearly integrated with the syntactic affinity matrices of the dependency trees to produce the enriched affinity matrices for GCNs in EFP. The extensive experiments show that the proposed model is very effective for EFP.

\section{Related Work}

EFP is one of the fundamental tasks in Information Extraction. The early work on this problem has employed the rule-based approaches \cite{nairn2006computing, sauri2008factuality, lotan2013truthteller} or the machine learning approaches (with manually designed features) \cite{diab2009committed, prabhakaran2010automatic, de2012did, lee2015event}, or the hybrid approaches of both \cite{sauri2012you, qian2015two}. Recently, deep learning has been applied to solve EFP. \cite{QianLZZZ18} employ Generative Adversarial Networks (GANs) for EFP while \cite{Rudinger} utilize LSTMs for both sequential and dependency representations of the input sentences. Finally, deep learning has also been considered for the related tasks of EFP, including event detection \cite{Nguyen:15b,Nguyen:16b,Lu:18,Nguyen:19}, event realis classification \cite{Mitamura:15,Nguyen:16g}, uncertainty detection \cite{adel2016exploring}, modal sense classification \cite{marasovic2016multilingual} and entity detection \cite{Nguyen:16d}. 

%stanovsky2017integrating
\section{Model}
%In this section, we describe the details of the proposed model. We first provide a formal definition of the task:

The formal definition of the EFP task is as follows. Let $(x_1,x_2,\ldots,x_n)$ be a sentence that contains some event mention of interest, where $n$ is the number of words/tokens and $x_i$ is the $i$-th token in the sentence. Also, let $k$ be the position of the anchor word in this sentence (i.e., token $x_k$). For EFP, the goal is to assign a real number in the range of [-3, +3] to quantify the degree to which the current event mention has happened. There are three major components in the EFP model proposed in this work, i.e., (i) sentence encoding, (ii) structure induction, and (iii) prediction.

%We use the range of [-3, +3] to follow the standards in the recent datasets for this problem \cite{lee2015event,Rudinger}.

\subsection{Sentence Encoding}

%accepts tokenized sentences to perform several layers of computation, each layer returns a matrix whose rows correspond to the contextualized embedding vectors of the words in the input sentences for the current layer. It has been shown that such contextualized word embeddings from BERT can help to

%, standing for Bidirectional Encoder Representations from Transformers,
% (i.e, question answering, natural language inference)

%The first step in most of the deep learning models for natural language processing (NLP) is to convert each word in the sentences into an embedding vector. 

The first step is to convert each word in the sentences into an embedding vector. In this work, we employ the contextualized word representations BERT in \cite{devlin2018bert} for this purpose. 
BERT is a pre-trained language representation model with multiple computation layers that has been shown to improve many NLP tasks. 
In particular, the sentence $(x_1,x_2,...,x_n)$ would be first fed into the pre-trained BERT model from which the contextualized embeddings of the words in the last layer are used for further computation. We denote such word embeddings for the words in $(x_1,x_2,\ldots,x_n)$ as $(e_1,e_2,\ldots,e_n)$ respectively.

%As $(e_1,e_2,\ldots,e_n)$ are general word representations, 
% (similar to \cite{Rudinger})
In the next step, we further abstract $(e_1,e_2,\ldots,e_n)$ for EFP by feeding them into two layers of bidirectional LSTMs (as in \cite{Rudinger}). This produces $(h_1,h_2,\ldots,h_n)$ as the hidden vector sequence in the last bidirectional LSTM layer (i.e., the second one). We consider $(h_1,h_2,\ldots,h_n)$ as a rich representation of the input sentence $(x_1,x_2,\ldots,x_n)$ where each vector $h_i$ encapsulates the context information of the whole input sentence with a greater focus on the current word $x_i$.

\subsection{Structure Induction}

%Given the hidden representation $(h_1,h_2,\ldots,h_n)$, it is possible to use the hidden vector corresponding to the anchor word $h_k$ as the features to perform factuality prediction (as done in \cite{Rudinger}). However, despite the rich context information over the whole sentence, the features in $h_k$ are not directly designed to focus on the context words that are important for the factuality prediction of the current event mention (i.e., the cue words). In order to explicitly encode the information of the cue words into the representations for the anchor word, we propose to learn an importance matrix $A = \left(a_{ij}\right)_{i,j=1..n}$ in which the value in the cell $a_{ij}$ quantifies the contribution of the context word $x_i$ for the hidden representation at $x_j$ if the representation vector at $x_j$ is used to form features for EFP. The importance matrix $A$ would then be used as the adjacent/weight matrix in the graph convolutional neural networks (GCNs) \cite{kipf2016semi} to accumulate the current hidden representations of the context words into the new hidden representations for each word in the sentence.

Given the hidden representation $(h_1,h_2,\ldots,h_n)$, it is possible to use the hidden vector corresponding to the anchor word $h_k$ as the features to perform factuality prediction (as done in \cite{Rudinger}). However, despite the rich context information over the whole sentence, the features in $h_k$ are not directly designed to focus on the import context words for factuality prediction. In order to explicitly encode the information of the cue words into the representations for the anchor word, we propose to learn an importance matrix $A = \left(a_{ij}\right)_{i,j=1..n}$ in which the value in the cell $a_{ij}$ quantifies the contribution of the context word $x_i$ for the hidden representation at $x_j$ if the representation vector at $x_j$ is used to form features for EFP. The importance matrix $A$ would then be used as the adjacent/weight matrix in the graph convolutional neural networks (GCNs) \cite{kipf2016semi,Nguyen:18} to accumulate the current hidden representations of the context words into the new hidden representations for each word in the sentence.

In order to learn the weight matrix $A$, as presented in the introduction, we propose to leverage both semantic and syntactic structures of the input sentence. In particular, for the semantic structure, we use the representation vectors from LSTMs for $x_i$ and $x_j$ (i.e., $h_i$ and $h_j$) as the features to compute the contribution score in the cell $a^{sem}_{ij}$ of the semantic weight matrix $A^{sem} = (a^{sem}_{ij})_{i,j=1..n}$:
\begin{equation*}
\begin{split}
h'_i &=  \text{tanh}(W^{sem}_1 h_i)\\
a^{sem}_{ij} &=  \text{sigmoid}(W^{sem}_2 [h'_i,h'_j])
\end{split}
\end{equation*}

%, $\sigma$ is the Sigmoid function, and $W^{sem}_1$ and $W^{sem}_2$ are model parameters.

Note that we omit the biases in the equations of this paper for convenience. In the equations above, $[h'_i,h'_j]$ is the concatenation of $h'_i$ and $h'_j$. Essentially, $a^{sem}_{ij}$ is a scalar to determine the amount of information that should be sent from the context word $x_i$ to the representation at $x_j$ based on the semantic relevance for EFP.

In the next step for the syntactic structure, we employ the dependency tree for the input sentence to generate the adjacent/weight matrix $A^{syn} = (a^{syn}_{ij})_{i,j=1..n}$, where $a^{syn}_{ij}$ is set to 1 if $x_i$ is connected to $x_j$ in the tree, and 0 otherwise. Note that we augment the dependency trees with the self-connection and reverse edges to improve the coverage of the weight matrix.

Finally, the weight matrix $A$ for GCNs would be the linear combination of the sematic structure $A^{sem}$ and the syntactic structure $A^{syn}$ with the trade-off $\lambda$: 
$$A = \lambda A_{sem} + (1-\lambda) A_{syn}$$
Given the weight matrix $A$, the GCNs \cite{kipf2016semi} are applied to augment the representations of the words in the input sentence with the contextual representations for EFP. In particular, let $H_0$ be the the matrix with $(h_1,h_2,\ldots,h_n)$ as the rows: $H_0 = [h_1,h_2,\ldots,h_n]$. One layer of GCNs would take an input matrix $H_i$ ($i \ge 0$) and produce the output matrix $H_{i+1}$: $H_{i+1} = g ( A H_i W^g_i )$ where $g$ is a non-linear function. In this work, we employ two layers of GCNs (optimized on the development datasets) on the input matrix $H_0$, resulting in the semantically and syntactically enriched matrix $H_2$ with the rows of $(h^g_1, h^g_2, \ldots, h^g_n)$ for EFP. 

\begin{table*}[t!]
%\small
%\addtolength{\abovecaptionskip}{-4.0mm}
%\addtolength{\belowcaptionskip}{-3.mm}
\begin{center}
\resizebox{.8\textwidth}{!}{
\begin{tabular}{lcccccccc}
  & \multicolumn{2}{c}{FactBank} & \multicolumn{2}{c}{UW} & \multicolumn{2}{c}{Meantime} & \multicolumn{2}{c}{UDS-IH2} \\
  & MAE & $r$ & MAE & $r$ & MAE & $r$ & MAE & $r$ \\
  \hline
  \cite{lee2015event}* & - & - & 0.511 & 0.708 & - & - & - & - \\
  \cite{stanovsky2017integrating}* & 0.590 & 0.710 & \bf 0.420 & 0.660 & 0.340 & 0.470 & - & - \\ 
  %\hline
 L-biLSTM(2)-S*\dag & 0.427 & 0.826 & 0.508 & 0.719 & 0.427 &  0.335 & 0.960 & 0.768 \\
  \hline
    L-biLSTM(2)-MultiBal**\dag & 0.391 & 0.821 & 0.496 & 0.724 &  0.278 & 0.613 & - & - \\
     L-biLSTM(1)-MultiFoc**\dag & 0.314 & 0.846 & 0.502 & 0.710 & 0.305 & 0.377 & - & - \\
     L-biLSTM(2)-MultiSimp w/UDS-IH2**\dag & 0.377 & 0.828 & 0.508 & 0.722 & 0.367 & 0.469 & 0.965 & 0.771 \\
     %\hline
      H-biLSTM(1)-MultiSimp**\dag & 0.313 & 0.857 & 0.528 & 0.704 & 0.314 &  0.545 & - & - \\
        H-biLSTM(2)-MultiSimp w/UDS-IH2**\dag & 0.393 & 0.820 & 0.481 & 0.749 & 0.374 &  0.495 & 0.969 & 0.760 \\
        \hline
        L-biLSTM(2)-S+BERT* & 0.381 & 0.85 & 0.475 & 0.752 & 0.389 & 0.394 & 0.895 & 0.804 \\
        L-biLSTM(2)-MultiSimp w/UDS-IH2+BERT** & 0.343 & 0.855 & 0.476 & 0.749 & 0.358 & 0.499 & 0.841 & 0.841 \\
        H-biLSTM(1)-MultiSimp+BERT** & 0.310 & 0.821 & 0.495 & 0.771 & 0.281 & 0.639 & 0.822 & 0.812 \\
        H-biLSTM(2)-MultiSimp w/UDS-IH2+BERT** & 0.330 & 0.871 & 0.460 & 0.798 & 0.339 & 0.571 & 0.835 & 0.802 \\
        \hline
  Graph-based (Ours)* & 0.315 & 0.890 & 0.451 & 0.828 & 0.350 & 0.452 & 0.730 & 0.905 \\
  Ours with multiple datasets** & \bf 0.310 & \bf 0.903 & 0.438 & \bf 0.830 & \bf 0.204 & \bf 0.702 & \bf 0.726 & \bf 0.909 \\
\end{tabular}
}
\end{center}
\caption{\label{results} Test set performance. * denotes the models trained on separate datasets while ** indicates those trained on multiple datasets. \dag specifies the models in \cite{Rudinger} that are significantly improved with BERT.
%\ddag specifies the models with BERT.
%The best performance in each column is in bold. 
  }
\end{table*}

\subsection{Prediction}

% In this component, we aggregate the context-aware representation vectors in $H_2$ to generate an overall representation of the input event mention for factuality prediction. In particular, we use the representation vector of the anchor word (i.e., $h^g_k$) as the query to compute the attention weights for each vector in $H_2$, from which the weighted sum is obtained to function as the feature vector $V$:

This component predicts the factuality degree of the input event mention based on the context-aware representation vectors $(h^g_1, h^g_2, \ldots, h^g_n)$. In particular, as the anchor word is located at the $k$-th position (i.e., the word $x_k$), we first use the vector $h^g_k$ as the query to compute the attention weights for each representation vector in $(h^g_1, h^g_2, \ldots, h^g_n)$. These attention weights would then be employed to obtain the weighted sum of $(h^g_1, h^g_2, \ldots, h^g_n)$ to produce the feature vector $V$:
\begin{equation*}
\begin{split}
 \alpha_i  &= W^a_1 h_k^g \cdot (W^a_2 h_i^g)^\top\\
 \alpha_1', \alpha_2', \ldots, \alpha_n'  &= \text{softmax}(\alpha_1, \alpha_2, \ldots, \alpha_n)\\
V  &= \sum_i \alpha_i' W^a_3 h^g_i
\end{split}
\end{equation*}
where $W^a_1$, $W^a_2$ and $W^a_3$ are the model parameters. 
The attention weights $\alpha_i'$ would help to promote the contribution of the important context words for the feature vector $V$ for EFP.

Finally, similar to \cite{Rudinger}, the feature vector $V$ is fed into a regression model with two layers of feed-forward networks to produce the factuality score. Following \cite{Rudinger}, we train the proposed model by optimizing the Huber loss with $\delta = 1$ and the Adam optimizer with learning rate = 1.0.

\section{Experiments}

%This section describes our experiments for EFP. For details of the datasets and model parameters see appendix \ref{dataset}.

\subsection{Datasets, Resources and Parameters}

\begin{table}[t!]
%\addtolength{\abovecaptionskip}{-3.0mm}
%\addtolength{\belowcaptionskip}{-3.5mm}
\centering
\small
\resizebox{0.5\textwidth}{!}{
\begin{tabular}{lcccc}
    \hline
  Dataset & \multicolumn{1}{c}{Train} & \multicolumn{1}{c}{Dev} & \multicolumn{1}{c}{Test} & \multicolumn{1}{c}{Total} \\
  \hline
  FactBank & 6636 & 2462 & 663 & 9761 \\
  MEANTIME & 967 & 210 & 218 & 1395 \\
  UW & 9422 & 3358 & 864 & 13644 \\
  UDS-IH2 & 22108 & 2642 & 2539 & 27289 \\ 
  \hline
\end{tabular}
}
\caption{\label{statistics} The numbers of examples in each dataset}
\end{table}

Following the previous work \cite{stanovsky2017integrating,Rudinger}, we evaluate the proposed EFP model using four benchmark datasets: FactBack \cite{sauri2009factbank}, UW \cite{lee2015event}, Meantime \cite{minard2016meantime} and UDS-IH2 \cite{Rudinger}. The first three datasets (i.e., FactBack, UW, and Meantime) are the unified versions described in \cite{stanovsky2017integrating} where the original annotations for these datasets are scaled to a number in [-3, +3]. For the fourth dataset (i.e., UDS-IH2), we follow the instructions in \cite{Rudinger} to scale the scores to the range of [-3, +3]. Each dataset comes with its own training data, test data and development data. Table \ref{statistics} shows the numbers of examples in all data splits for each dataset used in this paper.

We tune the parameters for the proposed model on the development datasets. The best values we find in the tuning process include: 300 for the number of hidden units in the bidirectional LSTM layers, 1024 for the dimension of the projected vector $h_i'$ in the structure induction component, 300 for the number of feature maps for the GCN layers, 600 for the dimention of the transformed vectors for attention based on ($W^a_1$, $W^a_2$, $W^a_3$), and 300 for the number of hidden units in the two layers of the final regression model. For the trade-off parameter $\lambda$ between the semantic and syntactic structures, the best value for the datasets FactBack, UW and Meantime is $\lambda = 0.6$ while this value for UDS-IH2 is $\lambda = 0.8$.

% Details about the datasets and parameters are presented in Appendix \ref{dataset}.

% Following the previous work \cite{stanovsky2017integrating,Rudinger}, we evaluate the EFP models in this work using four benchmark datasets: FactBack \cite{sauri2009factbank}, UW \cite{lee2015event}, Meantime \cite{minard2016meantime} and UDS-IH2 \cite{Rudinger}. The first three datasets (i.e., FactBack, UW, and Meantime) are the unified versions described in \cite{stanovsky2017integrating} where the original annotations for these datasets are scaled to a number between -3 and +3. For the fourth dataset (i.e., UDS-IH2), we follow the instructions described in \cite{Rudinger} to scale the scores to the range of [-3, +3]. Each dataset comes with its own training data, test data and development data.

% We tune the parameters for the proposed model on the development datasets. The best values we find in the tuning process include: 300 for the number of hidden units in the bidirectional LSTM layers, 1024 for the dimension of the projected vector $h_i'$ in the structure induction component, 300 for the number of feature maps for the GCN layers, 600 for the dimention of the transformed vectors for attention based on ($W^a_1$, $W^a_2$, $W^a_3$), and 300 for the number of hidden units in the two layers of the final regression model. For the trade-off parameter $\lambda$ between the semantic and syntactic structures, the best value for the datasets FactBack, UW and Meantime is $\lambda = 0.6$ while this value for UDS-IH2 is $\lambda = 0.8$.

\subsection{Comparing to the State of the Art}

\begin{table*}[t!]
%\addtolength{\abovecaptionskip}{-3.0mm}
%\addtolength{\belowcaptionskip}{-3.5mm}
% \centering
% % \resizebox{0.5\textwidth}{!}{
% \begin{tabular}{lcccc}
%   & \multicolumn{1}{c}{FactBank} & \multicolumn{1}{c}{UW} & \multicolumn{1}{c}{Meantime} & \multicolumn{1}{c}{UD-IH2} \\
%   \hline
%   The proposed model & \bf 0.903 & \bf 0.830 & \bf 0.702 & \bf 0.909 \\ \hline
%   - syntax structure ($\lambda=1$) & 0.867 & 0.801 & 0.658 & 0.893 \\
%   - semantic structure ($\lambda=0$) & 0.832 & 0.782 & 0.604 & 0.862 \\
%   - structure induction component & 0.821 & 0.735 & 0.582 & 0.828 \\
%   %LINEAR - Bert & 0.751 & 0.602 & 0.551 & 0.806 \\
%   %LINEAR - Attention & 0.823 & 0.715 & 0.612 & 0.820
%   - BERT & 0.831 & 0.751 & 0.570 & 0.817 \\
%   - attention in prediction & 0.890 & 0.821 & 0.695 & 0.899 \\
% \end{tabular}
% % }

\begin{center}
\resizebox{.8\textwidth}{!}{
\begin{tabular}{lcccccccc}
  & \multicolumn{2}{c}{FactBank} & \multicolumn{2}{c}{UW} & \multicolumn{2}{c}{Meantime} & \multicolumn{2}{c}{UDS-IH2} \\
  & MAE & $r$ & MAE & $r$ & MAE & $r$ & MAE & $r$ \\
  \hline
  The proposed model & \bf 0.310 &  \bf 0.903 & \bf 0.438 & \bf 0.830 & \bf 0.204 & \bf 0.702 & \bf 0.726 & \bf 0.909 \\
  \hline
 - syntax structure ($\lambda=1$) & 0.314 & 0.867 & 0.442 & 0.801 & 0.251 & 0.658 & 0.753 & 0.893 \\
    - semantic structure ($\lambda=0$) & 0.337 & 0.832 & 0.449 & 0.782 & 0.288 & 0.604 & 0.798 & 0.862 \\
     - structure induction component & 0.352 & 0.821 & 0.457 & 0.735 & 0.305 & 0.582 & 0.855 & 0.828 \\
     - BERT & 0.342 & 0.831 & 0.462 & 0.751 & 0.315 & 0.570 & 0.896 & 0.817 \\
     %\hline
      - attention in prediction & 0.312 & 0.890 & 0.441 & 0.821 & 0.221 & 0.695 & 0.737 & 0.899 \\
\end{tabular}
}
\end{center}

\caption{\label{ablation} Correlation ($r$) and MAE for different model configurations. The model without BERT (i.e., - BERT) uses Glove \cite{pennington2014glove} as in \cite{Rudinger}. 
  }
\end{table*}

This section evaluates the effectiveness of the proposed model for EFP on the benchmark datasets. We compare the proposed model with the best reported systems in the literature with linguistic features \cite{lee2015event,stanovsky2017integrating} and deep learning \cite{Rudinger}. Table \ref{results} shows the performance. Importantly, to achieve a fair comparison, we obtain the actual implementation of the current state-of-the-art EFP models from \cite{Rudinger}, introduce the BERT embeddings as the inputs for those models and compare them with the proposed models (i.e., the rows with ``+BERT''). Following the prior work, we use MAE (Mean Absolute Error), and $r$ (Pearson Correlation) as the performance measures.

In the table, we distinguish two methods to train the models investigated in the previous work: (i) training and evaluating the models on separate datasets (i.e., the rows associated with *), and (ii) training the models on the union of FactBank, UW and Meantime, resulting in single models to be evaluated on the separate datasets (i.e., the rows with **). It is also possible to train the models on the union of all the four datasets (i.e., FactBank, UW, Meantime and UDS-IH2) (corresponding to the rows with w/UDS-IH2 in the table). From the table, we can see that in the first method to train the models the proposed model is significantly better than all the previous models on FactBank, UW and UDS-IH2 (except for the MAE measure on UW), and achieves comparable performance with the best model \cite{stanovsky2017integrating} on Meantime. In fact, the proposed model trained on the separate datasets also significantly outperforms the current best models on FactBank, UW and UDS-IH2 when these models are trained on the union of the datasets with multi-task learning (except for MAE on Factbank where the performance is comparable). Regarding the second method with multiple datasets for training, the proposed model (only trained on the union of FactBank, UW and Meantime) is further improved, achieving better performance than all the other models in this setting for different datasets and performance measures. Overall, the proposed model yields the state-of-the-art performance over all the datasets and measures (except for MAE on UW with comparable performance), clearly demonstrating the advantages of the model in this work for EFP.

\subsection{Ablation Study}

Table \ref{ablation} presents the performance of the proposed model when different elements are excluded to evaluate their contribution. 
We only analyze the proposed model when it is trained with multiple datasets (i.e., FactBank, UW and Meantime). However, the same trends are observed for the models trained with separate datasets. As we can see from the table, both semantic and syntactic information are important for the proposed model as eliminating any of them would hurt the performance. Removing both elements (i.e., not using the structure induction component) would significantly downgrade the performance.
Finally, we see that both the BERT embeddings and the attention in the prediction are necessary for the proposed model to achieve good performance. 

%Moreover, from Table \ref{ablation} we can see BERT and structure induction component have almost equal contribution to the model performance. Our hypothesis is that contextual word embedding, e.g. BERT, can incorporate structural information earlier into the model. Thus, similar to structure induction component, BERT can improve the EFP performance. 
% Finally, we see that BERT is very helpful as it significantly improves the proposed model.

%Semantic structure is more effective as its removal causes greater performance loss.

%

\section{Conclusion \& Future Work}

We present a graph-based deep learning model for EFP that exploits both syntactic and semantic structures of the sentences to effectively model the important context words. We achieve the state-of-the-art performance over several EFP datasets. 

% Despite the benefit of the syntactic structures to complement the semantic information, 
One potential issue with the current approach is that it is dependent on the existence of the high-quality dependency parser. Unfortunately, such parser is not always available in different domains and languages. Consequently, in the future work, we plan to develop methods that can automatically induce the sentence structures for EFP.
% to avoid the reliance on the external dependency parser.

%The extensive experiments demonstrate the effectiveness of the proposed model, leading to the state-of-the-art performance over the benchmark datasets.

\subsection*{Acknowledgement} This research is partially supported by the NSF grant CNS-1747798 to the IUCRC Center for Big Learning. 

\bibliography{acl2019}

\begin{thebibliography}{28}
\expandafter\ifx\csname natexlab\endcsname\relax\def\natexlab#1{#1}\fi

\bibitem[{Adel and Sch{\"u}tze(2017)}]{adel2016exploring}
Heike Adel and Hinrich Sch{\"u}tze. 2017.
\newblock Exploring different dimensions of attention for uncertainty
  detection.
\newblock In \emph{Proceedings of the 15th Conference of the European Chapter
  of the Association for Computational Linguistics: Volume 1, Long Papers},
  pages 22--34.

\bibitem[{De~Marneffe et~al.(2012)De~Marneffe, Manning, and Potts}]{de2012did}
Marie-Catherine De~Marneffe, Christopher~D Manning, and Christopher Potts.
  2012.
\newblock Did it happen? the pragmatic complexity of veridicality assessment.
\newblock \emph{Computational linguistics}, 38(2):301--333.

\bibitem[{Devlin et~al.(2018)Devlin, Chang, Lee, and
  Toutanova}]{devlin2018bert}
Jacob Devlin, Ming-Wei Chang, Kenton Lee, and Kristina Toutanova. 2018.
\newblock Bert: Pre-training of deep bidirectional transformers for language
  understanding.
\newblock \emph{arXiv preprint arXiv:1810.04805}.

\bibitem[{Diab et~al.(2009)Diab, Levin, Mitamura, Rambow, Prabhakaran, and
  Guo}]{diab2009committed}
Mona~T Diab, Lori Levin, Teruko Mitamura, Owen Rambow, Vinodkumar Prabhakaran,
  and Weiwei Guo. 2009.
\newblock Committed belief annotation and tagging.
\newblock In \emph{Proceedings of the Third Linguistic Annotation Workshop},
  pages 68--73. Association for Computational Linguistics.

\bibitem[{Kipf and Welling(2016)}]{kipf2016semi}
Thomas~N Kipf and Max Welling. 2016.
\newblock Semi-supervised classification with graph convolutional networks.
\newblock \emph{arXiv preprint arXiv:1609.02907}.

\bibitem[{Lee et~al.(2015)Lee, Artzi, Choi, and Zettlemoyer}]{lee2015event}
Kenton Lee, Yoav Artzi, Yejin Choi, and Luke Zettlemoyer. 2015.
\newblock Event detection and factuality assessment with non-expert
  supervision.
\newblock In \emph{Proceedings of the 2015 Conference on Empirical Methods in
  Natural Language Processing}, pages 1643--1648.

\bibitem[{Lotan et~al.(2013)Lotan, Stern, and Dagan}]{lotan2013truthteller}
Amnon Lotan, Asher Stern, and Ido Dagan. 2013.
\newblock Truthteller: Annotating predicate truth.
\newblock In \emph{Proceedings of the 2013 Conference of the North American
  Chapter of the Association for Computational Linguistics: Human Language
  Technologies}, pages 752--757.

\bibitem[{Lu and Nguyen(2018)}]{Lu:18}
Weiyi Lu and Thien~Huu Nguyen. 2018.
\newblock Similar but not the same: Word sense disambiguation improves event
  detection via neural representation matching.
\newblock In \emph{Proceedings of the Conference on Empirical Methods in
  Natural Language Processing (EMNLP)}.

\bibitem[{Marasovic and Frank(2016)}]{marasovic2016multilingual}
Ana Marasovic and Anette Frank. 2016.
\newblock Multilingual modal sense classification using a convolutional neural
  network.
\newblock \emph{Proceedings of the 1st Workshop on Representation Learning for
  NLP, Rep4NLP@ACL 2016, Berlin, Germany, August 11, 2016}, pages 111--120.

\bibitem[{Minard et~al.(2016)Minard, Speranza, Urizar, Altuna, van Erp, Schoen,
  van Son et~al.}]{minard2016meantime}
A-L Minard, Manuela Speranza, Ruben Urizar, Begona Altuna, MGJ van Erp,
  AM~Schoen, CM~van Son, et~al. 2016.
\newblock Meantime, the newsreader multilingual event and time corpus.
\newblock In \emph{Proceedings of the Tenth International Conference on
  Language Resources and Evaluation (LREC 2016)}. Portoro{\v{z}}, Slovenia.

\bibitem[{Mitamura et~al.(2015)Mitamura, Liu, and Hovy}]{Mitamura:15}
Teruko Mitamura, Zhengzhong Liu, and Eduard Hovy. 2015.
\newblock Overview of tac kbp 2015 event nugget track.
\newblock In \emph{Proceedings of Text Analysis Conference (TAC)}.

\bibitem[{Nairn et~al.(2006)Nairn, Condoravdi, and
  Karttunen}]{nairn2006computing}
Rowan Nairn, Cleo Condoravdi, and Lauri Karttunen. 2006.
\newblock Computing relative polarity for textual inference.
\newblock In \emph{Proceedings of the fifth international workshop on inference
  in computational semantics (icos-5)}.

\bibitem[{Nguyen et~al.(2016a)Nguyen, Cho, and Grishman}]{Nguyen:16a}
Thien~Huu Nguyen, Kyunghyun Cho, and Ralph Grishman. 2016a.
\newblock Joint event extraction via recurrent neural networks.
\newblock In \emph{Proceedings of the Annual Conference of the North American
  Chapter of the Association for Computational Linguistics (NAACL)}.

\bibitem[{Nguyen et~al.(2016b)Nguyen, Fu, Cho, and Grishman}]{Nguyen:16b}
Thien~Huu Nguyen, Lisheng Fu, Kyunghyun Cho, and Ralph Grishman. 2016b.
\newblock A two-stage approach for extending event detection to new types via
  neural networks.
\newblock In \emph{Proceedings of the 1st ACL Workshop on Representation
  Learning for NLP (RepL4NLP)}.

\bibitem[{Nguyen and Grishman(2015b)}]{Nguyen:15b}
Thien~Huu Nguyen and Ralph Grishman. 2015b.
\newblock Event detection and domain adaptation with convolutional neural
  networks.
\newblock In \emph{Proceedings of the Annual Meeting of the Association for
  Computational Linguistics (ACL)}.

\bibitem[{Nguyen and Grishman(2018)}]{Nguyen:18}
Thien~Huu Nguyen and Ralph Grishman. 2018.
\newblock Graph convolutional networks with argument-aware pooling for event
  detection.
\newblock In \emph{Proceedings of the Association for the Advancement of
  Artificial Intelligence (AAAI)}.

\bibitem[{Nguyen et~al.(2016g)Nguyen, Meyers, and Grishman}]{Nguyen:16g}
Thien~Huu Nguyen, Adam Meyers, and Ralph Grishman. 2016g.
\newblock New york university 2016 system for kbp event nugget: A deep learning
  approach.
\newblock In \emph{Proceedings of Text Analysis Conference (TAC)}.

\bibitem[{Nguyen et~al.(2016d)Nguyen, Sil, Dinu, and Florian}]{Nguyen:16d}
Thien~Huu Nguyen, Avirup Sil, Georgiana Dinu, and Radu Florian. 2016d.
\newblock Toward mention detection robustness with recurrent neural networks.
\newblock In \emph{Proceedings of IJCAI Workshop on Deep Learning for
  Artificial Intelligence (DLAI)}.

\bibitem[{Nguyen and Nguyen(2019)}]{Nguyen:19}
Trung~Minh Nguyen and Thien~Huu Nguyen. 2019.
\newblock One for all: Neural joint modeling of entities and events.
\newblock In \emph{Proceedings of the Association for the Advancement of
  Artificial Intelligence (AAAI)}.

\bibitem[{Pennington et~al.(2014)Pennington, Socher, and
  Manning}]{pennington2014glove}
Jeffrey Pennington, Richard Socher, and Christopher Manning. 2014.
\newblock Glove: Global vectors for word representation.
\newblock In \emph{Proceedings of the Conference on Empirical Methods in
  Natural Language Processing (EMNLP)}, pages 1532--1543.

\bibitem[{Prabhakaran et~al.(2010)Prabhakaran, Rambow, and
  Diab}]{prabhakaran2010automatic}
Vinodkumar Prabhakaran, Owen Rambow, and Mona Diab. 2010.
\newblock Automatic committed belief tagging.
\newblock In \emph{Proceedings of the 23rd International Conference on
  Computational Linguistics: Posters}, pages 1014--1022. Association for
  Computational Linguistics.

\bibitem[{Qian et~al.(2018)Qian, Li, Zhang, Zhou, and Zhu}]{QianLZZZ18}
Zhong Qian, Peifeng Li, Yue Zhang, Guodong Zhou, and Qiaoming Zhu. 2018.
\newblock Event factuality identification via generative adversarial networks
  with auxiliary classification.
\newblock In \emph{Proceedings of the Twenty-Seventh International Joint
  Conference on Artificial Intelligence, {IJCAI} 2018, July 13-19, 2018,
  Stockholm, Sweden.}, pages 4293--4300.

\bibitem[{Qian et~al.(2015)Qian, Li, and Zhu}]{qian2015two}
Zhong Qian, Peifeng Li, and Qiaoming Zhu. 2015.
\newblock A two-step approach for event factuality identification.
\newblock In \emph{Asian Language Processing (IALP), 2015 International
  Conference on}, pages 103--106. IEEE.

\bibitem[{Rudinger et~al.(2018)Rudinger, White, and Van~Durme}]{Rudinger}
Rachel Rudinger, Aaron~Steven White, and Benjamin Van~Durme. 2018.
\newblock Neural models of factuality.
\newblock In \emph{Proceedings of the 2018 Conference of the North American
  Chapter of the Association for Computational Linguistics: Human Language
  Technologies, Volume 1 (Long Papers)}, pages 731--744.

\bibitem[{Saur{\'\i}(2008)}]{sauri2008factuality}
Roser Saur{\'\i}. 2008.
\newblock A factuality profiler for eventualities in text.
\newblock \emph{Unver{\"o}ffentlichte Dissertation, Brandeis University.
  Zugriff auf http://www. cs. brandeis. edu/\~{} roser/pubs/sauriDiss}, 1.

\bibitem[{Saur{\'\i} and Pustejovsky(2009)}]{sauri2009factbank}
Roser Saur{\'\i} and James Pustejovsky. 2009.
\newblock Factbank: a corpus annotated with event factuality.
\newblock \emph{Language resources and evaluation}, 43(3):227.

\bibitem[{Saur{\'\i} and Pustejovsky(2012)}]{sauri2012you}
Roser Saur{\'\i} and James Pustejovsky. 2012.
\newblock Are you sure that this happened? assessing the factuality degree of
  events in text.
\newblock \emph{Computational Linguistics}, 38(2):261--299.

\bibitem[{Stanovsky et~al.(2017)Stanovsky, Eckle-Kohler, Puzikov, Dagan, and
  Gurevych}]{stanovsky2017integrating}
Gabriel Stanovsky, Judith Eckle-Kohler, Yevgeniy Puzikov, Ido Dagan, and Iryna
  Gurevych. 2017.
\newblock Integrating deep linguistic features in factuality prediction over
  unified datasets.
\newblock In \emph{Proceedings of the 55th Annual Meeting of the Association
  for Computational Linguistics (Volume 2: Short Papers)}, volume~2, pages
  352--357.

\end{thebibliography}
\bibliographystyle{acl_natbib}

% \clearpage

% \appendix

% \section{Motivating Example}
% \label{parse-example}
% For the given example: ``{\it I will, after seeing the treatment of others, go back when I need medical care.}'', the dependency tree is shown in figure \ref{parse}.

% \section{Datasets and Parameters}
% \label{dataset}

%We used Dev set for fine tuning the parameters and test set to evaluate the performance on each dataset.

\end{document}